\documentclass[11pt]{article}

\usepackage{microtype}
\usepackage{graphicx}
\usepackage{booktabs} 
\usepackage{amssymb}
\usepackage{amsmath}
\usepackage[ruled, linesnumbered]{algorithm2e}
\usepackage{subcaption}
\usepackage{multirow}
\usepackage{mathtools}
\usepackage{siunitx}
\usepackage{color}
\usepackage{ulem}
\usepackage{sidecap}
\usepackage{wrapfig}
\usepackage{lipsum}

\graphicspath{ {./images/} }
\newcommand{\rt}{\right}
\newcommand{\lt}{\left}

\usepackage{microtype} 
\usepackage{booktabs}  

\usepackage{amsmath}
\usepackage{amsthm}

\usepackage[final]{automl}
\usepackage{natbib}

\title{Reinforced Meta Active Learning} 

\author[1]{\nameemail{Michael Katz}{michael.c.katz@gmail.com}}
\author[1]{\nameemail{Eli Kravchik}{eli.kravchik@huawei.com}}

\affil[1]{Huawei Technologies \\ Israel Research Center}

\hypersetup{%
  pdfauthor={M Katz, E Kravchik}, 
  pdftitle={Reinforced Meta Active Learning},
  pdfsubject={Reinforced Meta Active Learning}
}

\begin{document}


\abovedisplayskip=12pt plus 3pt minus 9pt
\belowdisplayskip=12pt plus 3pt minus 9pt
\abovedisplayshortskip=0pt plus 3pt
\belowdisplayshortskip=7pt plus 3pt minus 4pt

\maketitle

\begin{abstract}
In stream-based active learning, the learning procedure typically has access to a stream of unlabeled data instances and must decide for each instance whether to label it and use it for training or to discard it. 
There are numerous active learning strategies which try to minimize the number of labeled samples required for training in this setting by identifying and retaining the most informative data samples. 
Most of these schemes are rule-based and rely on the notion of \emph{uncertainty}, which captures how small the distance of a data sample is from the classifier's decision boundary. 
Recently, there have been some attempts to learn optimal selection strategies directly from the data, but many of them are still lacking generality for several reasons: 1) They focus on specific classification setups, 2) They rely on rule-based metrics, 3) They require offline pre-training of the active learner on related tasks. 
In this work we address the above limitations and present an online stream-based meta active learning method which learns on the fly an informativeness measure directly from the data, and is applicable to a general class of classification problems without any need for pretraining of the active learner on related tasks. 
The method is based on reinforcement learning and combines episodic policy search and a contextual bandits approach which are used to train the active learner in conjunction with training of the model. 
We demonstrate on several real datasets that this method learns to select training samples more efficiently than existing state-of-the-art methods.

\end{abstract}
\section{Introduction}
In many machine learning applications, data for training the model arrives in the form of a sequential stream of unlabeled data samples, where some or all of these samples are then labeled and used for training. Since annotated data is typically difficult and expensive to obtain, especially in domains where reliable labels can only be obtained by experts, active learning (AL) alleviates this difficulty by focusing the labeling budget only on informative data instances which enable to train the model most effectively. However, in some situations, due to practical limitations, e.g. storage constraints at edge devices, etc., upon the arrival of each data sample, an immediate decision must be made whether to retain the sample for training or to discard it. Such a setting is known as \emph{stream-based active learning}, as opposed to \emph{pool-based active learning}, where all the unlabeled data samples are available to the active learner right from the start.
Examples for such learning scenarios include spam filtering \citep{sculley2007online}, human activity recognition \citep{adaimi2019leveraging}, sentiment analysis \citep{smailovic2014stream}, learning from social-media \citep{pohl2018batch}, collecting data from fleet vehicles for training perception systems for self-driving cars, and more. In fact, this scenario is applicable to any learning system which accumulates large amounts of unlabeled data and wishes to use only a small representative subset for training, but is unable to store the data to be selected for labeling at a later time, and must decide on the spot whether to keep each data instance or not. 

Over the years, several rule-based approaches were proposed for the stream-based AL setting \citep{settles2009active,lewis1994sequential,seung1992query,sculley2007online,chu2011unbiased,zliobaite2014active,kottke2015probabilistic,derosa2017confidence}.  
Most of these approaches attempt to assign an informativeness measure to each data sample and give preference to samples with a high measure, also known as uncertainty sampling. Other approaches maintain a committee of classifiers and select data samples for which there is a large disagreement over their predictions by all committee members.
More recently, learning-based approaches which learn AL policies directly from the data were also reported \citep{woodward2017active,fang2017learning,kvistad2019augmented,wassermann2019ral,krawczyk2019adaptive,cheng2013feedback}. Most of these approaches try to harness reinforcement learning in order to learn optimal active learning policies for training a classifier. However, many of them either suffer from lack of generality or are impractical as an online algorithm for AL. For instance, \citep{woodward2017active} is applicable only in a specific and non-standard classification setup known as one-shot learning, \citep{wassermann2019ral,cheng2013feedback} are based on combinations of rule-based informativeness measures such as uncertainty sampling, and finally \citep{fang2017learning} requires pre-training of the active learner on a related task before applying it on the actual task of interest, rendering the approach impractical for any new task.

In this work we address the limitations of previously reported meta-learning approaches for AL from several perspectives. First, we do not limit the active learner to consider any specific function of the model's predictions for making a decision to select samples for training. Secondly, we consider a meta-learning strategy which is applicable to any classification problem, and finally, our approach is an online approach, which does not require any offline pre-training. Our main contributions are summarized as follows:

\begin{itemize}
	\item A novel stream-based meta AL method 
	which adapts itself to the task and dataset online and learns to select on the fly the most informative data samples for training a classifier (our method is also suitable for a regression type of problem).
	\item The method is based on reinforcement learning and combines episodic policy optimization and a contextual bandits (CB) approach which are applied \emph{online}. In particular, the method does not require any offline pre-training on related tasks and/or datasets.
	\item An empirical evaluation demonstrating that for a given budget constraint imposed by the system, this method learns to select training samples more efficiently than existing uncertainty-based approaches. 
\end{itemize}


\section{Related work} 
Most stream-based AL algorithms use rule-based query strategies to evaluate the informativeness of unlabeled data samples. 
The most commonly used query strategy is \emph{uncertainty sampling} \citep{lewis1994sequential}, where the active learner queries the instances for which it is least certain about their most probable label. The uncertainty can be defined in many ways, e.g., in terms of the maximum prediction value across classes, the difference between the two highest prediction values a.k.a. \emph{margin sampling}, and even the Shannon entropy of the prediction vector or the mutual information between predictions and model parameters \citep{gal2017deep}.
Another query strategy is \emph{query-by-committee} (QBC) \citep{seung1992query}, where a committee of models which are all trained on the labeled instances represent competing hypotheses. 
Each model member is allowed to vote on the labeling of query candidates and the candidates over which the committee members disagree most are considered more informative. 



Other rule-based approaches reported more recently include 
\citep{kottke2015probabilistic}, which defines a probabilistic spatial usefulness measure to model the true posterior probability of the classifier and uses it to select the best samples, \citep{fujii2016budgeted}
which applies pool-based AL approaches based on adaptive submodularity for the stream-based setting, and 
\citep{derosa2017confidence} which analyzes confidence intervals for splitting leaves in decision trees and uses them for selective sampling in online decision tree learning.



Some learning-based approaches aimed at learning selection strategies directly from the data were also recently reported.
For example, \citep{cheng2013feedback} proposes a reinforcement AL algorithm that adaptively mixes two sample selection criteria based on the KL divergence measured from classifier change.
In \citep{woodward2017active} meta learning and reinforcement learning are combined to learn an active learner for the one-shot classification problem \citep{santoro2016one}. 
In \citep{fang2017learning} reinforcement learning is also used to train a stream-based active learner agent to select samples for training a classifier on a Natural Language Processing task. In this work the agent is first trained to select samples for one task, and then transferred to select samples for a similar task. This work is further extended in \citep{kvistad2019augmented} by introducing a memory-augmented neural architecture.
RAL \citep{wassermann2019ral} is a stream-based AL framework modeled as a CB problem. A committee of expert classifiers, where each member has a decision weight, is used to vote on whether or not to label an incoming sample. Once a decision is made, a reward signal which measures the usefulness of the committee's decision is used to update the decision weights of the classifiers. A similar pure bandits approach is also reported in \citep{krawczyk2019adaptive}. 

\section{A Reinforced Meta Active Learning Framework}\label{sec: A Reinforced}

\subsection{Problem Formulation - Stream-based Active Learning}\label{subsec:Problem Formulation}
Let $\mathcal{D}_0 = \left\{\left(x_i^{0},y_i^{0}\right)\right\}_{i=1}^{N_0}$ be an initial training set composed of pairs $(x_i^0,y_i^0)$ of input feature vectors $x_i^0 \in \mathbb{R}^d$ and corresponding ground truth labels ${y_i^0 \in \left\{1, ..., C\right\}}$ independently drawn from some joint probablity distribution. 
Next, let $\mathcal{X} = \left\{x_1, x_2, ..., x_N \right\}$ be a data stream of $N$ additional input feature vectors $x_i \in \mathbb{R}^{d}$, and $\mathcal{Y} = \left\{y_1, y_2, ..., y_N\right\}$ be their corresponding \emph{unknown} ground truth labels $y_i \in \left\{1, ..., C\right\}$ randomly drawn from the same distribution. 
We further assume the existence of an oracle annotator $c(x)$ who is able to recover the class $y$ associated with a given input feature vector $x$.
The initial training set $\mathcal{D}_0$ and the data stream $\mathcal{X}$ along with the oracle $c(x)$ are used to train a classifier model $p\left(y|x;\theta\right)$ parameterized by $\theta$ 
with a given training procedure. 

\begin{figure}[t]
	\centering
	\includegraphics[width=1.0\textwidth]{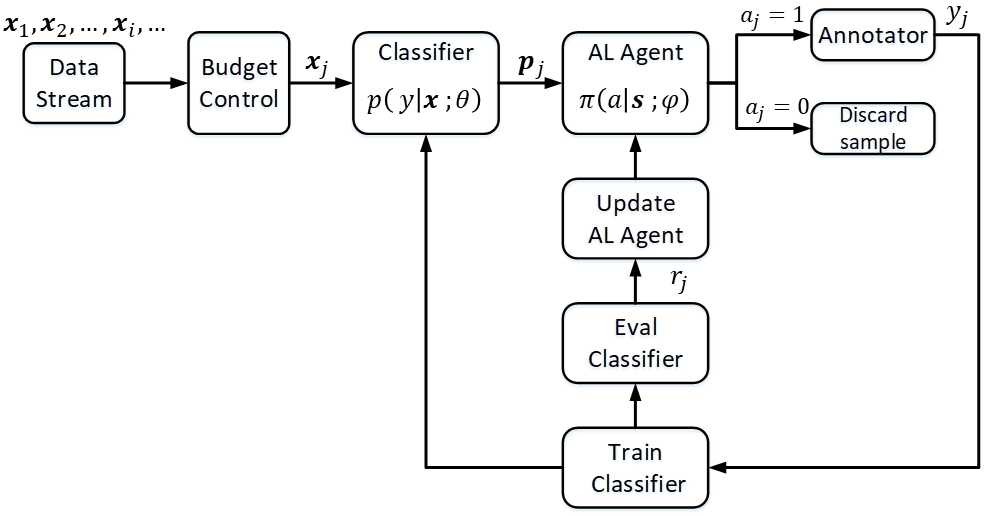}
	\caption{Framework of Reinforced Meta Active Learning}
	\label{fig:system}
\end{figure}

An active learner $\mathcal{L}$ for this setting is defined as follows.
At each time step $t$, the active learner computes a classifier $f_t \coloneqq p\left(y|x;\theta_t\right)$. 
At time step $t=0$, the classifier $f_0$ is obtained by training on the initial training set $\mathcal{D}_0$. 
Next, for $t = 1, 2, ..., N$, in response to a sample $x_t$ arriving from the stream, the active learner does one of the following actions: 1) Compute the label $y_t = c(x_t)$, add the pair $(x_t, y_t)$ to the training set, and re-train the classifier; or 2) Discard the sample, in which case the classifier remains unchanged. 
The active learner may utilize the classifier in order to make a decision. Denote by $N_u$ the total number of samples labeled by the active learner during the learning process until all $N$ samples from the stream are exhausted. 
Then, a budget constraint $b \in (0, 1]$ limits the fraction of total labeled samples such that $N_u \leq b N$ \citep{zliobaite2014active}.

Finally, let $\mathcal{U}_t$ be the training set accumulated by an active learner $\mathcal{L}$ up to time $t$. Then, the classifier at step $t$ is given by $f_{t} = \mathcal{L}\lt(f_{t-1}, \mathcal{U}_{t-1}, x_t \rt)$. Let $\textrm{Acc}(t)$ denote the accuracy of the classifier $f_t$ measured on a held out test set. Then, for a budget fraction $\beta \in (0, 1]$ we define the partial budget accuracy metric $\textrm{PBA}(\beta)$ as $\textrm{Acc}(t_{\beta})$, where $t_{\beta}$ is the smallest $t$ such that $\lt| \mathcal{U}_t\rt| \geq \beta b N$.

\subsection{System Setup}
The RMAL framework is illustrated in Figure \ref{fig:system}. Samples arrive in a stream to a budget controller, which monitors the number of samples $u_t$ selected for training until time $t$, and only passes on samples for which $u_t/t < b$. Samples for which this condition is not met are discarded. In order to prevent under consumption, whenever the instantaneous consumption rate $u_t/t$ falls below $b_{min}<b$, an amount of labeled samples are immediately replenished from the stream until the rate returns to a level of $b$.

A sample $x_j$ arriving from the budget controller is classified by the classifier, who produces a class prediction vector $\boldsymbol{p}_j \in [0,1]^C$. The prediction vector is then fed to an AL agent who makes a decision $a_j \in \{0, 1\}$  whether to select the current sample for training the classifier or to discard it. If the sample is chosen for training, the annotator produces the label $y_j$ and the pair $(x_j, y_j)$ is added to the training set, and the classifier is re-trained, otherwise the current step is done, and the system proceeds to process the next sample.

We take the approach of \citep{fang2017learning} and model the sequence of decisions, whether to label an input sample or not, as a Markov Decision Process (MDP), which allows the learning of a policy that can dynamically select instances most  informative for training. 
We represent the MDP as a tuple $\left(\mathcal{S}, \mathcal{A}, p\left(s_{j+1}|s_{j}, a_{j}\right), r(s, a)\right)$, where $\mathcal{S} \subseteq \mathbb{R}^{C}$ is the state space, $\mathcal{A}=\left\{0,1\right\}$ is the action space, $p\left(s_{j+1}|s_{j}, a_{j}\right)$ is the state transition probability, and $r(s, a)$ is the reward function. Under this framework, we represent the AL agent as a policy $\pi\left(a|s;\varphi\right)$ parameterized by $\varphi$. 
Once the classifier is re-trained, its accuracy is evaluated on a held out validation set, and used to produce a reward signal $r_j$ which is subsequently used to update the agent. This process repeats until all samples in the stream are exhausted.

In this framework there are two learning processes taking place: 1) learning of a classifier; and 2) learning of an agent. 
In the first process, the classifier is successively trained on an increasing set of samples chosen from the data stream by the agent. In the second process, the quality of the classifiers which were trained based on the agent's decisions is used to train the agent.
The interplay between these learning processes is further detailed in the next sections.

\section{An Alternating Approach}\label{sec:rmal_al}
Learning To Sample (LTS) \citep{shao2019learning} is a pool-based AL framework where a sampling model and a boosting model mutually learn from each other in iterations. 
The boosting model combines several weak models which are successively trained on a sequence of increasing datasets, and the sampling model is a regressor which estimates the uncertainty of unlabeled data samples. 

\IncMargin{1.5em}
\begin{algorithm}[t]
\SetKwInOut{KwIn}{Input}
\SetKwInOut{KwOut}{Output}
\SetKwInput{KwInit}{Initialization}
\SetKwFunction{TrainModel}{TrainModel}
\SetKwFunction{UpdateAgent}{UpdateAgent}
\SetKwFor{ForAll}{for all}{do}{endfall}
\SetKwComment{Comment}{$\triangleright$\ }{}
\SetArgSty{textnormal}
\Indm
\KwIn{Initial dataset $\mathcal{D}_0$, 
      Stream $\{x_1, ..., x_{N}\}$,
      Budget $b$, 
      Batch size $T$,
      Training episodes $E$\\
}
\KwOut{Classifier $f$}
\KwInit{$\pi \leftarrow \textrm{Random}$, $\mathcal{D}  \leftarrow \mathcal{D}_0$, $\mathcal{N} \leftarrow \emptyset$, $u \leftarrow 0$}
\Indp
\BlankLine
$f_0 \leftarrow \TrainModel\left(\mathcal{D}\right)$\\
$f^{\textrm{PXY}} \leftarrow f_0$ \\
Pretrain the policy $\pi$\\
\For{$t = 1, 2, ...,  N$}{
    $s_t \leftarrow f_{t-1}(x_t)$\\
    Draw action $a_t \sim \textrm{Bernoulli}(\pi(a=1|s_t))$\\
    \eIf{$(u/t<b)$ \textbf{and} $(a_t=1)$}{
        $y_t \leftarrow \textrm{query label of } x_t$\\
        $\mathcal{N}\leftarrow \mathcal{N} \cup \{(x_t, y_t)\}$\\ 
        $f_t \leftarrow \TrainModel(\mathcal{D} \cup \mathcal{N})$\\
        $u \leftarrow u + 1$\\
        \If{$|\mathcal{N}|=T$}{
            $\pi \leftarrow \UpdateAgent\left(\pi, f^{\textrm{PXY}}, \mathcal{D}, \mathcal{N}, E\right)$\\ 
            $\mathcal{D} \leftarrow \mathcal{D} \cup \mathcal{N}$\\
            $\mathcal{N} \leftarrow \emptyset$ \\
            $f^{\textrm{PXY}} \leftarrow f_t$
        }
    }{$f_t \leftarrow f_{t-1}$}
}\caption{RMAL-AL Algorithm} \label{alg:algorithm_rmal_al_1}
\end{algorithm}
\DecMargin{1.5em}

We borrow the alternating approach of LTS. The AL process continuously alternates between two interleaved learning phases: a classifier training phase and an agent training phase. In the classifier training phase, the classifier is trained and the agent remains fixed, while in the agent training phase, the classifier remains fixed and the agent is updated.   
A high level description of the algorithm is presented in Algorithm \ref{alg:algorithm_rmal_al_1}. We assume the existence of a training procedure $\texttt{TrainModel}(\mathcal{D})$ for training the model on a given dataset $\mathcal{D}$.
The AL process begins with training of the classifier on the initial training set $\mathcal{D}_0$ and initializing a proxy classifier (lines 1-2). The agent is then pretrained to mimic uncertainty sampling (we detail this procedure in section \ref{sec:Experimental Results}). Next, as long as there exists budget, each new sample is classified by the most up-to-date classifier, and the predictions are used by the agent to decide whether to use this sample for training. The decision is made stochastically by drawing from a Bernoulli distribution with parameter $\pi(a=1|s_t)$ (lines 5-7). Selected samples are annotated and added to the training set (lines 8-9), and a new classifier is trained on the new training set (line 10). Once an amount of $T$ samples has been accumulated, the agent undergoes an update phase where it is trained to select samples for training the proxy classifier (line 13).

\IncMargin{1.5em}
\begin{algorithm}[t]
\SetKwInOut{KwIn}{Input}
\SetKwInOut{KwOut}{Output}
\SetKwInput{KwInit}{Initialization}
\SetKwFunction{TrainModel}{TrainModel}
\SetKwFunction{Shuffle}{Shuffle}
\SetKwFunction{Accuracy}{Accuracy}
\SetKwComment{Comment}{$\triangleright$\ }{}
\SetArgSty{textnormal}
\Indm
\KwIn{Agent $\pi_{\varphi}$,
      Classifier $f$, 
      Training dataset $\mathcal{D}$,
      New labeled samples $\mathcal{N}=\{(z_1, y_1), ..., (z_{T}, y_{T})\}$,
      Number of episodes $E$,
      Validation dataset $\mathcal{V}$,\\
      Learning rate $\alpha$}
\KwOut{Updated agent $\pi_{\varphi}$}
\KwInit{$\mathcal{M} \leftarrow \emptyset$, $Acc_0 \leftarrow$ accuracy of $f$ on $\mathcal{V}$}
\Indp
\BlankLine
\For{episode = 1, 2, ..., $E$}{
    $f^{\textrm{PXY}}_0 \leftarrow f$ \\ 
    $\Shuffle(\mathcal{N})$\\
    \For{$t = 1, 2, ..., T$}{
        $s_t \leftarrow f^{\textrm{PXY}}_{t-1}(z_t)$\\
        Draw action $a_t \sim \textrm{Bernoulli}(\pi_{\varphi}(a=1|s_t)$\\
        \eIf{$a_t=1$}{
            $\mathcal{M}\leftarrow \mathcal{M} \cup \{(z_t, y_t)\}$\\
            $f^{\textrm{PXY}}_t \leftarrow \TrainModel(\mathcal{D} \cup \mathcal{M})$\\
            $\textrm{Acc}_t \leftarrow \Accuracy (f^{\textrm{PXY}}_t, \mathcal{V})$\\
            $r_t \leftarrow \left(\textrm{Acc}_t - \textrm{Acc}_{t-1}\right) / \textrm{Acc}_{t-1} $\\
        }{
            $f^{\textrm{CF}} \leftarrow \TrainModel(\mathcal{D} \cup \mathcal{M} \cup \{(z_t, y_t)\})$\\
            $\textrm{Acc}^{CF} \leftarrow \Accuracy (f^{\textrm{CF}}, \mathcal{V})$\\
            $r_t \leftarrow -(\textrm{Acc}^{CF} - \textrm{Acc}_{t-1}) / \textrm{Acc}_{t-1}$\\
            $f^{\textrm{PXY}}_t, \, \textrm{Acc}_t \leftarrow f^{\textrm{PXY}}_{t-1}, \, \textrm{Acc}_{t-1}$
        }
    }
    $\varphi \leftarrow \varphi + \alpha \nabla_{\varphi}J(\varphi)$ \Comment*[f]{Compute $ \nabla_{\varphi}J(\varphi)$ by eq. \ref{eq:rmal al policy gradient}}
}
\caption{RMAL-AL Algorithm: \texttt{UpdateAgent} (line 13 in Algorithm \ref{alg:algorithm_rmal_al_1})} \label{alg:algorithm_rmal_al_2}
\end{algorithm}
\DecMargin{1.5em}

The $\texttt{UpdateAgent}$ procedure, presented in Algorithm \ref{alg:algorithm_rmal_al_2}, is based on the REINFORCE policy search method \citep{williams1992simple}. We define a training episode of the agent as a sequence of $T$ steps, during which the agent selects samples for training a proxy classifier. The set $\mathcal{N}$ consists of the samples selected in the previous classifier training phase along with their labels. At the beginning of each episode, the proxy classifier $f^{\textrm{PXY}}_0$ is initialized to the state of the last classifier which was not yet trained on any of the samples in $\mathcal{N}$, and the set $\mathcal{N}$ is shuffled. 
Next, for each sample in $\mathcal{N}$, the proxy classifier is used to classify the sample, and the agent uses its prediction to decide stochastically whether or not to select the sample for training the proxy classifier. To facilitate policy learning we introduce an instantaneous reward signal which provides feedback on the quality of the actions taken by the agent.

In each step of the episode, if the agent decides to select the sample for training ($a=1$), the proxy classifier is re-trained with this sample\footnote{No annotation is required as this sample was already annotated in the previous training phase of the classifier.} and its accuracy on a held-out validation set is recorded as $\textrm{Acc}_t$. 
In this case, the reward signal $r_t$ is defined as the relative accuracy improvement of the model resulting from selecting the $t^{th}$ sample for training. 
If, however, the agent does not select the current sample for training, a mock classifier is trained with this sample and its validation accuracy is recorded as $\textrm{Acc}^{\textrm{CF}}$. We refer to this accuracy as counter-factual, as it is the  accuracy which would have been attained by the classifier, had the agent selected this sample for training. In this case, the reward signal $r_t$ is defined as minus the relative accuracy improvement of the model resulting from training on this samples. It follows that the instantaneous reward for a given state-action pair $(s_t, a_t)$ is given by:
\begin{equation} \label{eq:reward_rmal_al}
  r_t(s_t, a_t) =
    \begin{cases}
      \phantom{-}\left(\textrm{Acc}^{\phantom{\textrm{C}}}_t - \textrm{Acc}_{t-1}\right) / \textrm{Acc}_{t-1}, & a_t = 1\\
      -\left(\textrm{Acc}_t^{CF} - \textrm{Acc}_{t-1}\right) / \textrm{Acc}_{t-1}, & a_t = 0
    \end{cases}.       
\end{equation}

Our goal is to maximize the cumulative return $R(\tau) = \sum_{t=1}^{T}r_t$,
where $\tau$ is a trajectory of states, actions, and rewards $(s_1, a_1, r_1,  s_2, a_2, r_2, ...)$ occurring in one episode.
Thus, we wish to maximize the expected return
\begin{equation} \label{eq:pg_objective_rmal_al}
J(\varphi) = \mathbb{E}_{\tau \sim \pi_{\varphi}} \left[ R (\tau)\right]. 
\end{equation}
We use the REINFORCE method \citep{williams1992simple,sutton2018reinforcement} to update $\varphi$ via the gradient
\begin{equation} \label{eq:gradient_rmal_al}
    \nabla_{\varphi}J(\varphi) = \mathbb{E}_{\tau \sim \pi_{\varphi}} 
    \left[ \sum_{t=1}^{T} \nabla_{\varphi} \log \pi_{\varphi}(a_t|s_t) R(\tau)\right]
\end{equation}
which can be empirically approximated by sampling $m$ trajectories following policy $\pi_{\varphi}$ and averaging the result
\begin{equation} \label{eq:gradient_rmal_al_approx}
    \nabla_{\varphi}J(\varphi) \approx
    \frac{1}{m}\sum_{k=1}^{m}\sum_{t=1}^{T} \nabla_{\varphi} \log \pi_{\varphi}(a_t|s_t) 
    \left(R(\tau)-b(s_t)\right).
\end{equation}
$b(s_t)$ is any baseline function which estimates the expected return in state $s_t$, and does not depend on the actions. It is used to reduce the variance of the gradient's estimation.
Finally, we replace the return $R(\tau)$ in \ref{eq:gradient_rmal_al_approx} with the discounted future return $R_t = \sum_{t'=t}^{T}\gamma^{t'-t}r_{t'}$, where $\gamma \in [0, 1]$ is a discount rate which controls how much influence on future rewards we associate to each decision \citep{sutton2018reinforcement}. 
We discuss considerations for setting $\gamma$ in section \ref{sec:Experimental Results}. 
The gradient then becomes
\begin{equation} \label{eq:rmal al policy gradient}
    \nabla_{\varphi}J(\varphi) \approx
    \frac{1}{m}\sum_{k=1}^{m}\sum_{t=1}^{T} \nabla_{\varphi} \log \pi_{\varphi}(a_t|s_t) 
    \left(R_t-b(s_t)\right)
\end{equation}
and is used in line 19 of Algorithm \ref{alg:algorithm_rmal_al_2} to update the agent's weights. 

\section{A Hybrid Reinforcement Learning Approach}\label{sec:RMAL-HY}
In this section we incorporate into our meta AL framework a CB interpretation where the agent updates can be made simpler. We first discuss a pure CB solution 
and then describe how to combine it with our previous RMAL-AL algorithm.

\subsection{A Pure Contextual Bandits Strategy}
In contrast to a full reinforcement learning problem, in a CB setting, the distribution of the next state does not depend on the action taken by the agent. Thus, while taking an action in any state, the agent need not take delayed rewards into consideration and does not need to consider a long episode, rather a single state-action-reward $(s,a,r)$ triplet forms an episode, and the agent just focuses on selecting an optimal action for the given state.

One could argue that the assumption of independence between the actions and the next state does not hold in our AL setting, since the agent's actions indirectly affect the model which is trained on the samples selected by the agent, and the model affects the predictions for the next data sample which constitute the next state. 
However, the state in our case is largely dependent on the sample itself which is the input for the model's predictions, and the data sequence is i.i.d. So, the CB approach is a reasonable approximation which focuses the agent on selecting samples which give a maximal immediate accuracy improvement.
Besides reducing complexity, another motivation for pursuing this direction is the fact that by continuously updating the agent online, we are less prone to overfitting on a particular sample batch. 

We now give the system in Figure \ref{fig:system} a CB interpretation and describe how we can train the AL agent. We focus on the differences with respect to the RMAL-AL scheme.
One crucial difference in the contxtual bandits approach is that 
there is no distinction between a model training phase and an agent training phase, rather the two occur concurrently. Whereas in the RMAL-AL scheme, the agent was trained by running episodes on a batch of already labeled samples, here there is no notion of episode, rather, the agent is continuously updated based on the actual reward which comes from the previous agent's decisions.
Thus, the reward signal used to train the agent does not include the counter factual component, and is simply defined as $r_t(s_t, a_t) = \lt(\textrm{Acc}_t - \textrm{Acc}_{t-1}\rt) / \textrm{Acc}_{t-1}$. 

Next, instead of the objective in \ref{eq:pg_objective_rmal_al}, we wish to maximize the expected instantaneous reward $r_t(s_t, a_t)$, so our objective function becomes 
\begin{equation} \label{eq:pg_objective_rmal_cb}
J(\varphi) = \mathbb{E}_{s, a \sim \pi_{\varphi}} \left[ r(s, a)\right]
\end{equation}
and its policy gradient can be approximated by
\begin{equation} \label{eq:rmal cb policy gradient}
    \nabla_{\varphi}J(\varphi_t)
    \approx
    \frac{1}{m}\sum_{k=1}^{m} \nabla_{\varphi} \log \pi_{\varphi}(a_k|s_k) 
    \left(r_k-b(s_k)\right)
\end{equation}
where we average the gradient 
over $m$ successive labeled samples which were used to train the classifier. After each $m$ data samples are labeled and used to train the classifier, we update the agent using the gradient approximation in \ref{eq:rmal cb policy gradient}. 
The parameter $m$ is a hyper parameter.

\IncMargin{1.5em}
\begin{algorithm}[t]
\SetKwInOut{KwIn}{Input}
\SetKwInOut{KwOut}{Output}
\SetKwInput{KwInit}{Initialization}
\SetKwFunction{TrainModel}{TrainModel}
\SetKwFunction{UpdateAgent}{UpdateAgent}
\SetKwFor{ForAll}{for all}{do}{endfall}
\SetKwComment{Comment}{$\triangleright$\ }{}
\SetArgSty{textnormal}
\Indm
\KwIn{Initial dataset $\mathcal{D}_0$,
      Stream $\{x_1, ..., x_{N}\}$,
      Budget $b$,
      Validation set $\mathcal{V}$,
      Batch sizes $\left\{T_1, T_2, ...\right\}$, 
      Training episodes $\left\{E_1, E_2, ...\right\}$
}
\KwOut{Classifier $f$}
\KwInit{$\pi_{\varphi} \leftarrow \textrm{Random}$, $\mathcal{D}  \leftarrow \mathcal{D}_0$, $\mathcal{N} \leftarrow \emptyset$, $u \leftarrow 0$, $k \leftarrow 1$}
\Indp
\BlankLine
$f^{\textrm{PXY}} \leftarrow f_0 \leftarrow \TrainModel(\mathcal{D})$\\
Pretrain the policy $\pi$\\
\For{$t = 1, 2, ...,  N$}{
    $s_t \leftarrow f_{t-1}(x_t)$\\
    Draw action $a_t \sim \textrm{Bernoulli}(\pi_{\varphi}(a=1|s_t))$\\
    \eIf{$(u/t<b)$ \textbf{and} $(a_t=1)$}{
        $y_t \leftarrow \textrm{query label of } x_t$\\
        $\mathcal{N}\leftarrow \mathcal{N} \cup \{(x_t, y_t)\}$\\
        $f_t \leftarrow \TrainModel(\mathcal{D} \cup \mathcal{N})$\\
        $u \leftarrow u + 1$\\
        \eIf(\Comment*[f]{Contextual bandits updates}){$E_k = 0$}{
            $\textrm{Acc}_t \leftarrow $ accuracy of $f_t$ on $\mathcal{V}$\\
            $r_t \leftarrow \lt(\textrm{Acc}_t - \textrm{Acc}_{t-1}\rt) / \textrm{Acc}_{t-1}$\\
            \If{$|\mathcal{N}|>T_k$}{
                $\varphi \leftarrow \varphi + \alpha \nabla_{\varphi}J(\varphi)$ \Comment*{Compute $ \nabla_{\varphi}J(\varphi)$ by eq. \ref{eq:rmal cb policy gradient}}
                $\mathcal{D} \leftarrow \mathcal{D} \cup \mathcal{N}$, \,
                $\mathcal{N} \leftarrow \emptyset$, \,
                $k \leftarrow k + 1$
            }
        }(\Comment*[f]{Episodic RL updates}){ 
            \If{$|\mathcal{N}|>T_k$}{
                $\pi_{\varphi} \leftarrow \UpdateAgent\left(\pi_{\varphi}, f^{\textrm{PXY}}, \mathcal{D}, \mathcal{N}, E_k\right)$\\
                $\mathcal{D} \leftarrow \mathcal{D} \cup \mathcal{N}$, \,
                $\mathcal{N} \leftarrow \emptyset$, \,
                $k \leftarrow k + 1$, \,
                $f^{\textrm{PXY}} \leftarrow f_t$
            }
        }
    }{$f_t \leftarrow f_{t-1}$}
}
\caption{RMAL-HY Algorithm} \label{alg:algorithm_rmal_hy}
\end{algorithm}
\DecMargin{1.5em}

\subsection{Combining Contextual Bandits with RMAL-AL} \label{subsec:RMAL-HY-old}

We now incorporate the CB approach 
by allowing each batch of selected samples to be used for updating the agent either as was done in RMAL-AL by the procedure $\texttt{UpdateAgent}$, or as in the pure CB approach described above.

In this way, it is possible to balance between two complementary update strategies. The former can better fit the agent to the current model state and data statistics but requires more computation and is more prone to overfitting, while the later is less computationally demanding and has better chances of generalizing to the remainder of the data stream but its updates are more noisy. 

To better trade-of complexity and policy accuracy, we allow to choose the size of each sample batch and the number of episodes in that batch. 
Specifically, the sequences $\{T_i\}$ and $\{E_i\}$, $i=1, 2, ...$ define the size of
and the number of episodes in each batch, respectively, where a value of $E_i > 0$ means that at the end of batch $i$, the agent is updated as in the RMAL-AL scheme, i.e., based on rewards associated with a proxy classifier using $E_i$ episodes, while a value of ${E_i=0}$ means that the agent is updated as in the CB scheme, i.e., based on rewards computed directly on the classifier which is actively learned.
Thus, earlier batches can be longer with more training episodes, yielding a more accurate policy, while later batches can be shorter with fewer episodes allowing quicker policy updates. 
This hybrid scheme named RMAL-HY is illustrated in Algorithm \ref{alg:algorithm_rmal_hy}.

\section{Experimental Results} \label{sec:Experimental Results}
We evaluate our algorithms by comparing their performance against two baselines: (1) random sampling (RND) and (2) variable uncertainty (VU) proposed in \citep{zliobaite2014active} which is considered state-of-the-art for stream-based AL.
We experiment with one synthetic dataset (described below) and 5 real datasets extracted from the UCI Machine Learning Repository and kaggle whose details are listed in Figure \ref{fig:Real dataset table}.

We divide each dataset into four disjoint parts: an initial training set, a streaming part, a validation set, and a test set. In each experiment we use the dataset to train a logistic regressor. The agent's architecture consists of a fully connected network with 4 hidden layers of 256, 512, 256, and 256 neurons, a non-linearity of tanh in the 1\textsuperscript{st} layer, ReLU non-linearities in the remaining layers, and finally a Sigmoid layer at the output. For training the agent we use SGD with weight decay of \num{5e-4}. The discount rate $\gamma$ is set close to zero\footnote{We noticed little performance improvement by considering delayed rewards.}, $m$ is set to 1, the baseline $b(s_t)$ is an exponential moving average, and $b_{min}$ is set in the range 0.6-0.9 depending on the value of $b$. For each algorithm, we first train the model on the initial training set, and then let 
the AL algorithm select samples for training the classifier. The validation set is used by the RMAL algorithms to produce a reward signal for updating the agent as described in sections \ref{sec:rmal_al} and \ref{sec:RMAL-HY}. 

In the case of the real datasets, after training the model on the initial training set, and prior to processing the data stream, we perform pretraining of the agent's policy to mimic a policy based on uncertainty sampling \citep{zliobaite2014active}. The purpose of this procedure is to let the agent start with a policy which is known to perform well, and try to improve on it, rather than start learning an optimal policy from scratch. To this end, we generate a dataset of random probability vectors and train the agent in a supervised manner to produce a positive decision if the maximum element of the input probabililty vector is below a threshold ${T_{unc}=0.6}$. In the synthetic dataset, we do not perform such pretraining of the agent.


\subsection{A Synthetic Dataset}
In order to demonstrate the effectiveness of our approach in finding good policies for training the classifier, we test our scheme on a synthetic dataset which is specifically tailored such that the VU baseline attains a significant gain w.r.t. the RND baseline. Our target in this test is to show that our RMAL-HY scheme can learn a policy which attains at least as good a performance as the VU baseline even without pretraining of the agent. 

\begin{figure}[t]
\begin{subfigure}{0.385\textwidth}
\includegraphics[width=0.95\linewidth, 
                 trim=0em 0em 0em 0em, clip=true]
                 {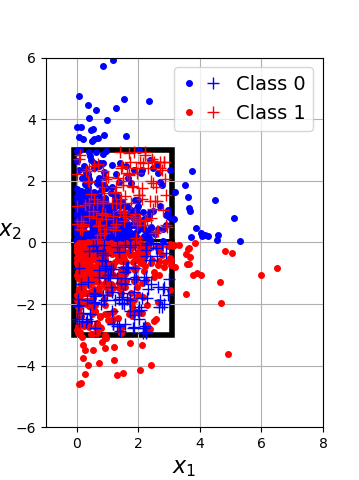}
\caption{}
\label{fig:exp_syn_dataset}
\end{subfigure}
\begin{subfigure}{0.615\textwidth}
\includegraphics[width=0.95\linewidth, 
                 trim=1.1em 0em 4em 0em, clip=true]
                 {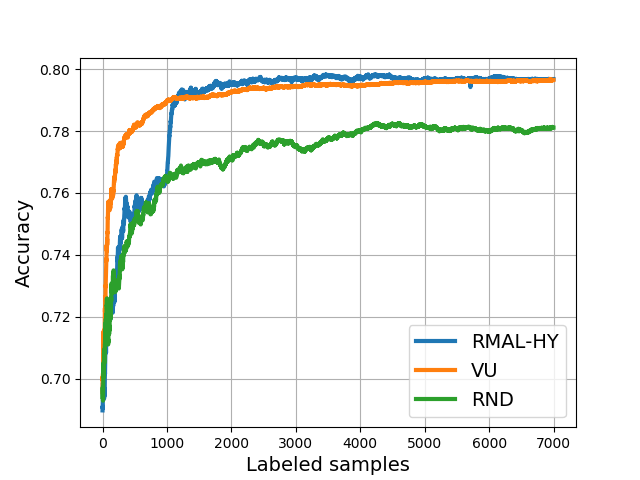}
\caption{}
\label{fig:rmal_hy_exponential_logistic_acc}
\end{subfigure}
\caption{(a) A two-dimensional synthetic dataset for which the VU baseline achieves superior performance to RND by a large margin. (b) Results of AL algorithms used for training a logistic regressor on the dataset shown in (a).}
\end{figure}

The synthetic dataset is a combination of an exponential distribution and a uniform one and is defined as follows. For a dataset of size $N$ and a constant $\alpha \in \left(0, \frac{1}{2}\right)$ we draw $\frac{(1-\alpha) N}{2}$ \mbox{2-dimensional} vectors from the exponential density $p(x_1, x_2) = e^{-x_1-x_2}, \, (x_1, x_2) \in [0, \infty]^2$, assign to them a label $y=0$, and then draw another $\frac{(1-\alpha) N}{2}$ 2-dimensional vectors from the density $p(x_1, x_2) = e^{-x_1+x_2}, \, (x_1, x_2) \in [0, \infty] \times [-\infty, 0]$, and assign to them a label $y=1$. Next we draw $\frac{\alpha N}{2}$ samples uniformly distributed in the rectangle $[0, 3] \times [0, 3]$, label them with $y=1$, and finally draw $\frac{\alpha N}{2}$ samples uniformly distributed in the rectangle $[0, 3] \times [-3, 0]$, and label them with $y=0$. For this dataset, it can be shown that an optimal decision boundary is the line $x_2 = 0$, and its accuracy would be equal to $1-\alpha$. A sample of this data set is plotted in Fig. \ref{fig:exp_syn_dataset} for $N=1200$ and $\alpha=0.2$. 

We use RMAL-HY to train a logistic regressor on this dataset in the stream-based AL setup described in section \ref{sec: A Reinforced}.
Fig. \ref{fig:rmal_hy_exponential_logistic_acc} shows the learning curve performance for a budget $b=0.1$.
The large performance gap between the RND and VU baselines is a result of this particular dataset. This is explained as follows. 
When the RND baseline is employed, in any given set of training samples, a fraction $\alpha$ of the samples causes the decision boundary to fluctuate due to the noisy labels.

However, in the VU baseline, by choosing samples closer to the decision boundary, the relative proportion of the outlier samples is dramatically reduced, leading to improved AL performance.
As is seen in Figure \ref{fig:rmal_hy_exponential_logistic_acc}, after some learning period, the RMAL-HY algorithm is able to learn an improved selection policy which is competitive with both baselines. Note that during the selection of the 1st batch of 1000 samples, the agent is still in its random initial state, so its performance coincides with the RND baseline. However, at the end of this batch, the 1st training phase of the agent is carried out, and the resulting policy learns to select more informative samples for training the classifier. In subsequent batches, the cheap online updates are enough to maintain the advantage of the policy even as more data continues to arrive.

\begin{figure}[t]
    \begin{minipage}[t]{0.51\textwidth}
        \vspace{-19pt}
        \includegraphics[width=0.99\linewidth, 
                 trim=0em 0em 0em 0em, clip=true]
                 {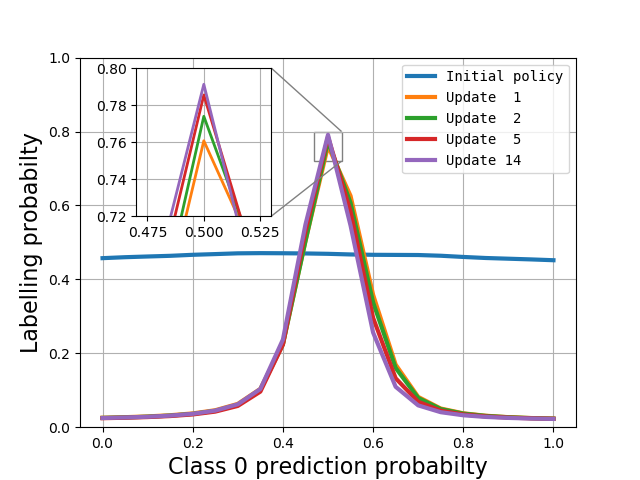}
        \captionsetup{width=0.9\linewidth}
        \caption{Evolution of policy updates exhibiting an increasing preference of uncertain data samples.}
        
        \label{fig:rmal_hy_exponential_logistic_policies}
    \end{minipage}
    \hfill
    \begin{minipage}[t]{0.48\textwidth}
        \vspace{0pt}
        \centering
        \renewcommand\arraystretch{1.4}
        \begin{tabular}[t]{|>{\raggedright}p{0.25\textwidth}
                        |>{\centering}p{0.18\textwidth}
                        |>{\centering}p{0.16\textwidth}
                        |>{\centering\arraybackslash}p{0.16\textwidth}|} 
        \hline
         \textbf{Dataset} & \textbf{Features} & \textbf{Classes} & \textbf{Samples}\\
         \noalign{\hrule height 1.0pt}
         Magic & 10 & 2 & 19,020\\
         \hline
         Activity & 8 & 4 & 75,128\\
         \hline
         Covertype & 54 & 7 & 581,012\footnotemark\\ 
         \hline
         Phishing & 30 & 2 & 11,054\\
         \hline
         Web Club Acceptance & 23 & 2 & 20,000\\
         \hline
        \end{tabular}
        \vfill
        \captionsetup{width=0.9\linewidth}
        \vspace*{20pt}
        \caption{A list of real datasets used for evaluating the RMAL-AL and RMAL-HY algorithms.}
        \label{fig:Real dataset table}
    \end{minipage}
\end{figure}

Figure \ref{fig:rmal_hy_exponential_logistic_policies} illustrates how the policies evolve during the AL process. Starting with the blue curve, which corresponds to a randomly initialized agent, these policies adapt to the underlying data distribution and progress towards more and more peaky distributions, which give increasing preference to data samples with prediction values closer to 1/2, i.e., samples with a higher level of uncertainty. Each curve is an average of corresponding policy updates in 100 trials.


\subsection{Real Datasets}

To further demonstrate the applicability of our scheme, we evaluate our algorithms on 5 real datasets with various number of attributes and classes, detailed in Figure \ref{fig:Real dataset table}. We perform our experiments for low budget values, e.g., 5\%, 10\%, 20\%, which are typical for an AL setting. For each budget value, we consider several values of $\beta$, e.g., 25\%, 50\%, and 100\%, in order to evaluate the performance of the algorithm during various stages of the AL process. We repeat each experiment for 20-100 trials, depending on the size of the dataset.

\begin{figure}[t]
    \centering 
\begin{subfigure}{0.41\textwidth}
  \includegraphics[width=\linewidth,
                   trim=0em 0em 0em 0em, clip=true]
                  {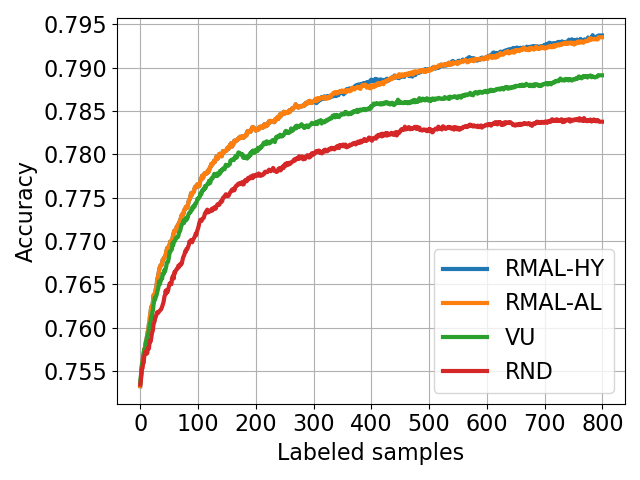}
  \caption{Magic}
  \label{fig:a}
\end{subfigure}\hfill\hfil 
\begin{subfigure}{0.4\textwidth}
  \includegraphics[width=\linewidth,
                   trim=0em 0em 0em 0em, clip=true]
                  {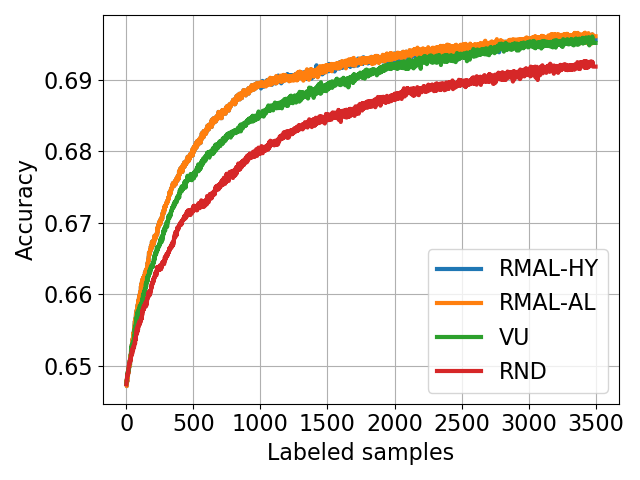}
  \caption{Covertype}
  \label{fig:b}
\end{subfigure}\hfil
\vspace{0.0cm}
\begin{subfigure}{0.41\textwidth}
  \includegraphics[width=\linewidth,
                   trim=0em 0em 0em 0em, clip=true]
                  {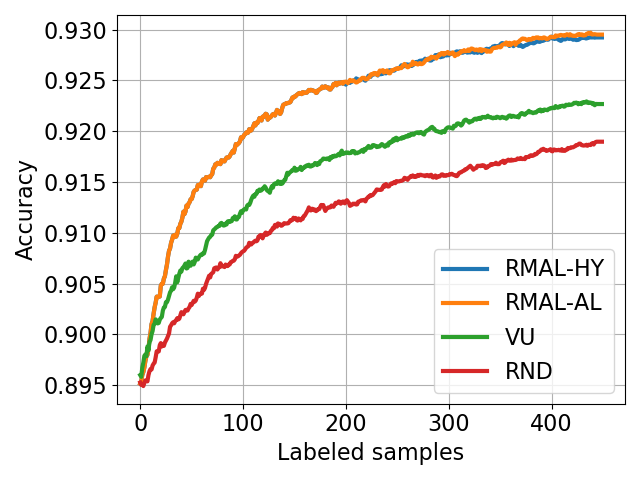}
  \caption{Phishing}
  \label{fig:c}
\end{subfigure}\hfill\hfil 
\begin{subfigure}{0.41\textwidth}
  \includegraphics[width=\linewidth,
                   trim=0em 0em 0em 0em, clip=true]
                  {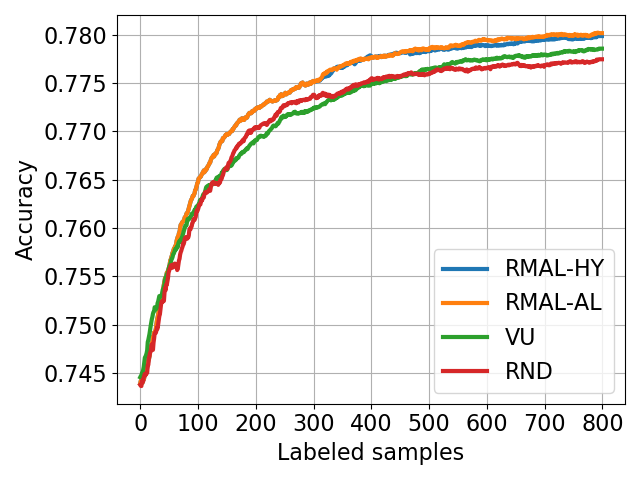}
  \caption{Web Club}
  \label{fig:d}
\end{subfigure}
\caption{Results of AL algorithms used for training a logistic regressor on several real datasets.}
\label{fig:real datasets learning curves}
\end{figure}

Figure \ref{fig:real datasets learning curves} shows learning curves for all considered AL algorithms when used to train a logistic regressor on 4 real datasets with a budget $b=0.05$. As can be seen, for very low budget values, our method outperforms the baselines throughout the active learning. 

Accuracy results of all experiments are summarized in Table \ref{tbl:Results}, were we report the average partial budget accuracy metric $\textrm{PBA}(\beta)$, defined in section \ref{subsec:Problem Formulation}. The largest accuracy for each setting is marked with boldface letters.
In all experiments, RMAL-AL was run with 50 episodes in the 1\textsuperscript{st} batch, 20 episodes in the 2\textsuperscript{nd} batch, and 10 episodes for the remaining batches with a learning rate of $0.1$. In RMAL-HY there were typically 100 episodes in the 1\textsuperscript{st} batch.
Results show a consistent advantage of the RMAL algorithms over both baselines across all considered datasets, across various budget values, and for different stages in the AL process.  

\footnotetext{In our experiments we use a subset of size 80,000.}

\begin{table}[b!]
    \centering
    \scriptsize
    \renewcommand\arraystretch{1.5}
    \begin{tabular}{|p{0.075\textwidth}
                    |>{\centering}p{0.035\textwidth}
                    !{\vrule width 1pt}>{\centering}p{0.04\textwidth}
                    |>{\centering}p{0.04\textwidth}
                    |>{\centering}p{0.04\textwidth}
                    !{\vrule width 1pt}>{\centering}p{0.04\textwidth}
                    |>{\centering}p{0.04\textwidth}
                    |>{\centering}p{0.04\textwidth}
                    !{\vrule width 1pt}>{\centering}p{0.04\textwidth}
                    |>{\centering}p{0.04\textwidth}
                    |>{\centering}p{0.04\textwidth}
                    !{\vrule width 1pt}>{\centering}p{0.04\textwidth}
                    |>{\centering}p{0.04\textwidth}
                    |>{\centering\arraybackslash}p{0.04\textwidth}|} 
    \hline
     \multirow{3}{*}{Dataset} & \multirow{3}{=}{$b [\%]$} & 
     \multicolumn{3}{c!{\vrule width 1pt}}{RMAL-HY} & \multicolumn{3}{c!{\vrule width 1pt}}{RMAL-AL} & \multicolumn{3}{c!{\vrule width 1pt}}{VU} & \multicolumn{3}{c|}{RND} \\
     \cline{3-14}
     & & \multicolumn{3}{c!{\vrule width 1pt}}{$\beta [\%]$} & \multicolumn{3}{c!{\vrule width 1pt}}{$\beta [\%]$} & \multicolumn{3}{c!{\vrule width 1pt}}{$\beta [\%]$} & \multicolumn{3}{c|}{$\beta [\%]$} \\
     \cline{3-14}
     & & \scriptsize{25} & \scriptsize{50} & \scriptsize{100} & \scriptsize{25} & \scriptsize{50} & \scriptsize{100} & \scriptsize{25} & \scriptsize{50} & \scriptsize{100} & \scriptsize{25} & \scriptsize{50} & \scriptsize{100} \\
     \noalign{\hrule height 1.0pt}
     \multirow{3}{*}{Magic} 
     & 5 & \textbf{78.27} & \textbf{78.86} & \textbf{79.37} & 	\textbf{78.27} & 78.78 & 79.35 &	78.05 & 78.54 & 78.91 &	77.76 & 78.18 & 78.38\\
     \cline{2-14}
     & 10 & \textbf{78.78} & \textbf{79.27} & 79.60 & 78.76 & \textbf{79.27} & \textbf{79.64} &	78.58 & 78.93 & 79.37 &	78.15 & 78.34 & 78.55\\
     \cline{2-14}
     & 20 & \textbf{79.13} & 79.56 & 79.71 &	79.11 & \textbf{79.61} & \textbf{79.72} &	78.96 & 79.37 & 79.47 &	78.38 & 78.46 & 78.69\\
     \hline
     \multirow{3}{*}{Activity} 
     & 5 & \textbf{89.68} & 90.18 & 90.88 & 89.67 & \textbf{90.19} & \textbf{90.95} & 88.94 & 89.95 & 90.61 & 83.73 & 87.81 & 89.71\\
     \cline{2-14}
     & 10 & \textbf{90.13} & 90.85 & 91.54 & 90.10 & \textbf{90.92} & \textbf{91.96} &	89.83 & 90.49 & 91.52 &	87.77 & 89.61 & 90.70\\
     \cline{2-14}
     & 20 & 90.73 & 91.13 & \textbf{91.93} &	\textbf{90.74} & \textbf{91.82} & 91.88 &	90.56 & 91.54 & 91.81 &	89.64 & 90.65 & 91.76\\
     \hline
     \multirow{3}{*}{Covertype} 
     & 5 & \textbf{68.81} & \textbf{69.25} & 69.56 & \textbf{68.81} & 69.23 & \textbf{69.61} & 68.36 & 69.08 & 69.52 &	67.83 & 68.60 & 69.18\\
     \cline{2-14}
     & 10 & \textbf{69.31} & \textbf{69.75} & \textbf{70.12} & \textbf{69.31} & 69.74 & 70.07 &	69.06 & 69.57 & 69.94 &	68.69 & 69.41 & 69.86\\
     \cline{2-14}
     & 20 & \textbf{69.68} & 70.00 & 70.77 & 69.65 & 70.15 & \textbf{70.97} & 69.63 & \textbf{70.17} & 70.88 & 69.21 & 69.93 & 70.75\\
     \hline
     \multirow{3}{*}{Phishing} 
     & 5 & \textbf{92.06} & 92.55 & 92.92 & \textbf{92.06} & \textbf{92.56} & \textbf{92.95} & 91.38 & 91.84 & 92.27 & 90.92 & 91.37 & 91.90\\
     \cline{2-14}
     & 10 & \textbf{92.33} & 92.70 & 92.82 & \textbf{92.33} & \textbf{92.78} & \textbf{92.85} &	91.88 & 92.41 & 92.68 & 91.34 & 91.81 & 92.28\\
     \cline{2-14}
     & 20 & 92.41 & 92.70 & 92.88 &	92.39 & \textbf{92.72} & \textbf{92.90} &	\textbf{92.44} & 92.65 & 92.75 & 91.88 & 92.33 & 92.55\\
     \hline
     \multirow{3}{=}{Web Club Acceptance} 
     & 5 & \textbf{77.23} & \textbf{77.78} & 77.98 & \textbf{77.23} & 77.76 & \textbf{78.01} & 76.91 & 77.50 & 77.85 & 77.04 & 77.55 & 77.74\\
     \cline{2-14}
     & 10 & \textbf{77.55} & \textbf{77.85} & \textbf{78.05} & \textbf{77.55} & 77.82 & \textbf{78.05} &	77.49 & 77.83 & 78.04 & 77.43 & 77.83 & 78.00\\
     \cline{2-14}
     & 20 & \textbf{77.91} & 77.97 & 78.04 &	77.89 & \textbf{78.00} & 78.06 &	77.80 & \textbf{78.00} & \textbf{78.09} & 77.70 & 77.88 & 77.98\\
     \hline
    \end{tabular}
    \caption{Performance comparison on real datasets.}
    \label{tbl:Results}
\end{table}

\section{Conclusions and Future Work}
In this work we present novel data-driven AL methods for selecting samples for training a classifier in a stream-based setting. 
The proposed methods, which combine several notions of reinforcement learning, learn to optimally select samples based on their model's predictions without any apriori rule-based functional restrictions or assumptions. 
Furthermore, these methods are applicable to a wide range of classification problems and are applied online without any need for offline training on any related task. 
We demonstrate on both a synthetic dataset and real datasets the efficacy of these methods with respect to state-of-the-art.

Although not pursued in this work, streaming data may sometimes exhibit changes in the underlying data distribution over time also known as concept drift. One future research direction would be to extend the meta active learner to adapt to these changes during the active learning process. For example, one could adapt the reward mechanism to account for changing statistics and thereby induce an appropriate change in the active learner. 

Another direction would be to alleviate the need to retrain the classifier after every selected sample. Allowing to train the model in batches may open the door for higher dimensional and large scale datasets and tasks, e.g,. image datasets which are typically trained in batches.


%
%
%
\bibliographystyle{apalike}
\bibliography{references}

\end{document}